\documentclass{egpubl}
\usepackage{mlvis2023}

 \WsPaper           

 \electronicVersion 

\ifpdf \usepackage[pdftex]{graphicx} \pdfcompresslevel=9
\else \usepackage[dvips]{graphicx} \fi

\PrintedOrElectronic

\usepackage{t1enc,dfadobe}

\usepackage{egweblnk}
\usepackage{cite}

\title[Interactive dense pixel visualizations for time series and model attribution explanations]%
      {Interactive dense pixel visualizations for \\time series and model attribution explanations}

\author[U. Schlegel \& D. Keim]
{\parbox{\textwidth}{\centering U. Schlegel$^{1}$\orcid{0000-0002-8266-0162}, and
D. A. Keim$^{1}$\orcid{0000-0001-7966-9740}
        }
        \\
{\parbox{\textwidth}{\centering $^1$University of Konstanz, Germany\\
      }
}
}

\begin{document}


\maketitle
\begin{abstract}
The field of Explainable Artificial Intelligence (XAI) for Deep Neural Network models has developed significantly, offering numerous techniques to extract explanations from models. 
However, evaluating explanations is often not trivial, and differences in applied metrics can be subtle, especially with non-intelligible data.
Thus, there is a need for visualizations tailored to explore explanations for domains with such data, e.g., time series.
We propose DAVOTS, an interactive visual analytics approach to explore raw time series data, activations of neural networks, and attributions in a dense-pixel visualization to gain insights into the data, models' decisions, and explanations.
To further support users in exploring large datasets, we apply clustering approaches to the visualized data domains to highlight groups and present ordering strategies for individual and combined data exploration to facilitate finding patterns.
We visualize a CNN trained on the FordA dataset to demonstrate the approach.


\end{abstract}  
\vspace{-3em}
\section{Introduction}
\the\font

Deep learning models slowly move into the time series field to boost the state-of-the-art performance and remove the tedious task of feature engineering~\cite{theissler_explainable_2022}.
However, due to the non-intelligible property of time series, adding another layer of complexity and a hard-to-understand model on top does not increase the interpretability of the process in general~\cite{spinner_explainer_2019}.
Thus, explainable AI (XAI) explores the extraction of explanations for the inner decision-making of complex models principally~\cite{guidotti_survey_2018}.
XAI and visualization methods help to understand single and global decisions for all kinds of models independently from the complexity~\cite{guidotti_survey_2018,spinner_explainer_2019}.
Some of these techniques can also be applied to deep learning models for time series data to extract explanations and support users to understand predictions~\cite{theissler_explainable_2022}.

However, most explanations are more accessible to understand and evaluate for their faithfulness if the underlying data is intelligible~\cite{theissler_explainable_2022}.
For non-intelligible data, such as time series, the evaluation can still be done similarly to intelligible data, e.g., images.
But judging faithfulness is more complicated due to only slight differences in the scores, and a visual assessment is effortful without heavy domain knowledge~\cite{schlegel_towards_2019}.
Thus, model users need enhanced data representation and explanations to grasp the model decision~\cite{schlegel_time_2021}.
There are many different approaches to representing time-oriented data~\cite{aigner_visualization_2011}.
However, combining raw time series with attributions is not trivial to visualize and needs some special care to support users in understanding the information and patterns hidden in the data~\cite{schlegel_time_2021,jeyakumar_how_2020}.

We propose a dense-pixel attribution visualization on time series (DAVOTS) to mitigate the challenges of non-intelligible data and attributions by breaking down all available information onto single pixels for users to explore.
Large time series datasets can be visualized with DAVOTS by showing univariate time series, activations, and attributions for a single sample as pixels in one row.
We extend these dense-pixel displays with histograms of the respective data.
To support users in exploring the data, our approach applies clustering methods to order the data and to enable users to find patterns and similar samples.
We focus on hierarchical clustering using three distance metrics and let users decide which clustering to explore.
As a use case, we trained a convolutional neural network on the FordA dataset~\cite{dau_ucr_2018} and explored it using DAVOTS.

Thus, our contributions are:
(1) a dense pixel visualization (DAVOTS) for raw time series, activations, and attributions of a model in the same row as pixel display,
(2) clustering methods using various distance functions and hierarchical clustering in the interactive visualization application DAVOTS,
(3) preliminary findings regarding patterns in the clustered data on a trained model on a time series dataset.
Based on related work~\cite{schlegel_time_2021,keim_designing_2000}, we gather challenges with current visualization techniques for attributions on time series on single samples and whole datasets.
Next, we describe how to overcome such challenges using clustering.
Afterward, we introduce the components of the DAVOTS application.
At last, we explore a CNN trained on the FordA dataset using DAVOTS.
\footnote{Running Demo at: {\tiny\href{https://davots.time-series-xai.dbvis.de/}{davots.time-series-xai.dbvis.de}}}

\section{Related Work}

Time series visualization provides various methods to present and highlight parts of time-oriented data~\cite{aigner_visualization_2011}.
In many cases, line plots are starting points for further analysis.
However, such a visualization technique does not scale well to large datasets or long time series~\cite{aigner_visualization_2011}.
Other techniques, such as small multiples or dense pixel displays, can overcome such challenges~\cite{aigner_visualization_2011}.
To support users in analyzing whole datasets and additional information, we need to incorporate, e.g., dense pixel displays.
However, the selected technique needs considerations to ensure user understanding, such as an understandable overview, details already in the overview, and filter options to recognize patterns~\cite{shneiderman_eyes_1996}.

Explainable AI (XAI) for time series has shown that attribution methods can be applied to any deep learning model~\cite{schlegel_towards_2019} and, in some cases, such as LIME~\cite{ribeiro_why_2016} and KernelSHAP~\cite{lundberg_unified_2017} also for every model~\cite{schlegel_towards_2019}.
However, visualizing these attributions can be tricky as it introduces another layer of information~\cite{schlegel_towards_2019}.
Schlegel and Keim~\cite{schlegel_time_2021} collected and reviewed currently available examples of possible visualizations for time series and attributions.
The time series line plot is often enhanced with a heatmap-like visualization.
As these visualizations do not use available space efficiently and are not scaleable for whole datasets, Assaf and Schumann~\cite{assaf_explainable_2019} use a purely pixel-based visualization to visualize the attributions but neglect the raw time series data.
Our technique also focuses on such a visualization, but we extend Assaf and Schumann~\cite{assaf_explainable_2019} approach to raw time series and scale the shown time series to whole datasets with as many samples as pixels on the screen.

To enable users to analyze the data further, we include clustering approaches to facilitate such an analysis of a whole dataset.
Time series clustering can be divided into three categories by the taxonomy of Aghabozorgi et al.~\cite{aghabozorgi_time_2015}, namely whole-time series, subsequence time series, and time point clustering.
We focused on the single time points as our visualizations present these to recognize patterns in the data further; thus, a time point clustering can help reveal such patterns.
However, there are other possible clustering approaches using subsequence time series, e.g., the shape of the time series to cluster the samples~\cite{aghabozorgi_time_2015}, which can also help to order the visualization approach in another way.

Different approaches combine time series clustering with visualization to enhance both methods. 
Van Wijk and van Selow~\cite{vanwijk_cluster_1999} use a normalized Euclidean distance with hierarchical clustering to find and visualize patterns in the power consumption of employees in an office.
They can recognize different events in the data more efficiently through such a combination.

\section{Clustering the data}

To enable users to identify patterns in the data, we cluster the raw data and the data produced by the model to get an ordering for our visualization.
Thus, we apply various time point clustering methods~\cite{aghabozorgi_time_2015} as we use time points for our visualization and want users to inspect the clustering patterns on the time points in the visualization.
Further, through our clustering, we can order the data based on the cluster results to inspect the data in groupings and enable visual patterns in the visualization.

We implement three different distance functions based on related work and surveys to apply clustering algorithms to our data.
To enable exploration of our data, we implement the Euclidean distance as a general baseline distance, a normalized Euclidean distance based on van Wijk and van Selow~\cite{vanwijk_cluster_1999}, and the Pearson correlation coefficient. 
The normalized Euclidean distance by van Wijk and van Selow~\cite{vanwijk_cluster_1999} normalizes both time series with their maximum and averages on the number of time points to improve the distance of similar shapes.
We selected these based on the literature~\cite{aghabozorgi_time_2015}, but we want to extend the functionality of our approach with different distances in future work.

As we do not always know the density distribution or possible partitions, we apply hierarchical agglomerative clustering on our data to have a flexible technique for distances and clusters.
Further, hierarchical clustering can also be used to order our input data through the merging order when points get merged into clusters.
Thus, enabling us to visualize the data in an order based on the clustering result.
Our current hierarchical clustering on the time points leads to initial orderings with which we can start the first analysis and apply our dense pixel visualization.

Our current approach implements the hierarchical clustering strategies by Ward~\cite{wardjr_hierarchical_1963} and complete linkage.
The approach by Ward~\cite{wardjr_hierarchical_1963} minimizes the variance in the data to agglomerate the data into clusters.
The complete linkage technique uses the maximum distance between clusters to merge these.
We also applied single and average linkage to the data in the first experiments to get clusters.
We decided on Ward and complete linkage based on the proposed measure of Guo and Gahegan~\cite{guo_spatial_2006}, as these lead to the best result using the Guo and Gahegan~\cite{guo_spatial_2006} measure.

The measure by Guo and Gahegan~\cite{guo_spatial_2006} analyzes the neighbors around the found clusters and calculates their distances.
As we want to have similar data vectors in our ordering near to each other to make patterns in the data more obvious, this measure looked promising.
However, we found out that most of the applied hierarchical clustering techniques do not differ much in the measure.
For future work, we want to extend the measure to more than one distance metric to include also the shape or other features.


\begin{figure*}[h!t]
  \centering
  \includegraphics[trim={0cm 1.7cm 0cm 0cm},clip,width=\linewidth]{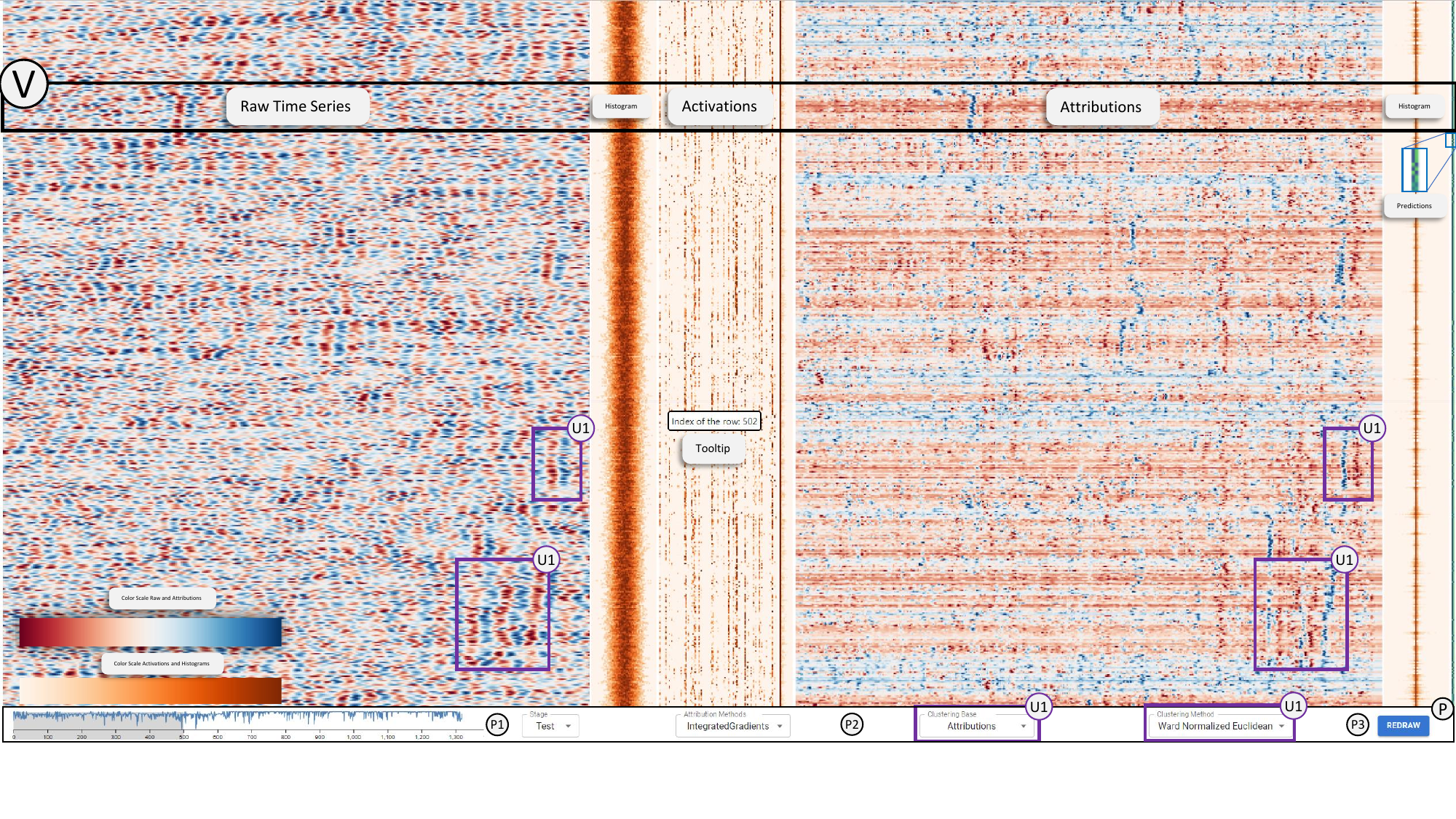}
  \caption{Overview of DAVOTS. (V) shows the visualization approach, while (P) shows the adjustable parameters. Starting from the left (V), the pixel visualization of the raw time series, next the histogram of the raw time series, afterward the activations of the model, then the corresponding histogram, next the attributions, then the corresponding histogram, and lastly, the probabilities of the prediction. On the bottom at (P), the parameters can be changed. (P1) demonstrates the standard deviation of every dataset sample for the selected clustering base as a slider to select the data which should be shown. (P2) presents the options for selecting the data and clustering. Here the stage (Train, Test, ...), the attribution methods, the clustering base, and the clustering method can be selected. With (P3), the complete visualization can be redrawn to incorporate the new parameters.}
  \label{fig:overview}
  \vspace{-2em}
\end{figure*}

\section{Dense-Pixel Attribution Visualization On Time Series}
Our dense-pixel attribution visualization on time series (DAVOTS) approach consists of a visualization and a parameter steering component.
Both have various interaction possibilities to steer the analysis in a particular direction.
We will start with the components and then introduce our current interactions.

\noindent\textbf{Components --}
Clustering the data enables users to recognize patterns in visualizations easier.
However, the challenge of visualizing a whole dataset and additional information still holds.
To overcome the issue, we propose to use a dense pixel visualization for our time series, activation, and attribution data.
Through such a visualization, we can present users with more information and use the space more efficiently for possible patterns.
We use as much available space as possible for our pixels for our visualization (\autoref{fig:overview} (V)).
The only margin we spare is the parameters on the bottom for the visualization (\autoref{fig:overview} (P)).
Thus, the whole visualization scales on the pixels of the device's resolution.

Our pixel visualization uses a tabular way to visualize the data.
Rows represent samples, and columns represent data groups, such as raw, attributions, or activation data.
So, every row from the top holds the complete data for one sample.
The row consists of the raw time series data (time from left to right), the histogram for the values of the raw time series, the activations of a selected linear layer, the histogram of these activations, the attributions for the input to the model, the histogram of these attributions, and lastly the prediction as can be seen in~\autoref{fig:overview} (V).
The corresponding histograms give a short overview of the distribution of the values in the data.
Each data type is color-coded to visualize its value on the pixel to fulfill the requirements and needs regarding the specific data type.

Due to the diverse data types having different properties, we need to select the color codes according to our needs on the data.
For time series, we use a diverging color scale as, in most cases, we are most interested in high and low values to find patterns.
The activations are shown using a sequential color scale starting with white for ReLUs and a diverging color scale with white in the middle for the sigmoid function.
For the attributions, we are also visualized using a diverging color scale with white in the middle to present high and low relevances in the attributions more quickly to the user.
All histograms use a sequential color scale starting from white to highlight the peak values again.
The last few pixels correspond to the prediction of the classifier layer of the model and use a diverging color scale with a different color than white in the middle to show the whole range of possible values without a focus.

Beneath the visualization (\autoref{fig:overview} (P), the options can be changed to modify the visualization.
A slider changes the amount of visualized samples and can be set by users, but defaults to a hundred samples at startup.
The slider is a brush on top of a line plot to facilitate the selection of samples that are more relevant for users and show some meta-information about the ordering.
We use the standard deviation on the sample level of the currently selected data and for the current ordering to highlight meta patterns and to introduce a high-level view of the samples.
However, other values, such as the mean, are also possible to use to present additional meta information.
Next, users can change the stage of the data towards the model.
E.g., training, testing, or validation.
After the stage, users can select the data for the ordering base data corresponding to the clustering.
For instance, the raw time series or the attributions can be selected.
Users can then select the clustering method based on the selected data to change the visualization.
E.g., an ordering based on the hierarchical clustering with the Ward method and the Euclidean distance.

\begin{figure*}[h!t]
    \centering
    \includegraphics[trim={0cm 9.8cm 0cm 0cm},clip,width=\linewidth]{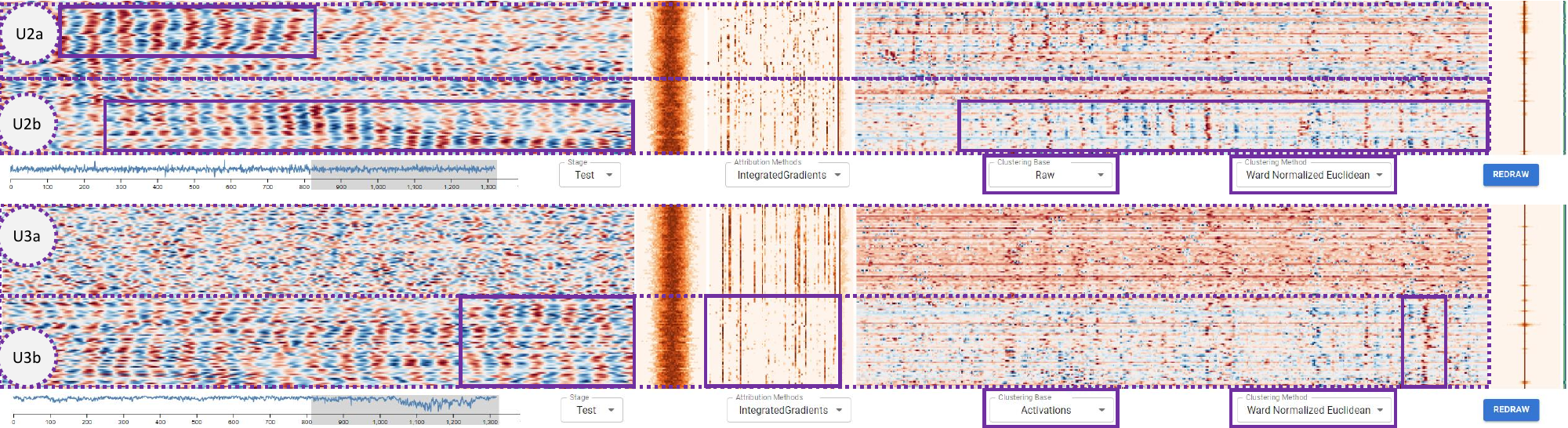}
    \caption{Two findings we have during our initial exploration on a CNN trained for the FordA dataset. (U2) presents the data clustered by the raw time series using the Ward method~\cite{wardjr_hierarchical_1963} with the normalized Euclidean distance~\cite{vanwijk_cluster_1999}. (U2a) presents on the left patterns in the raw time series, which are less obvious to observe in the attributions on the right. (U2b) shows clusters in the raw time series and smaller ones in the attributions. (U3) visualizes the clustering using the same settings as before on the activations. (U3a) shows mostly no patterns in the raw time series and attributions. (U3b) presents patterns on the activations, the raw time series, and also on the attributions.}
    \label{fig:finding}
    \vspace{-2em}
  \end{figure*}

\noindent\textbf{Interactions --}
Our current approach only focuses on representation and thus has only limited interactions.
The interaction limits itself to the parameter selection in which the line plot of the slider changes based on the changed settings.
Further, the possible applied clustering methods change on the selection of the base data as clustering on some data types does not make sense, e.g., probabilities of the prediction.
However, a small tooltip on top of the row, under the cursor, displays the index identifier of the sample (\autoref{fig:overview}) to enable users to find certain indices or extract indices for other applications.
A small highlight around the row under the cursor helps further to distinguish the hovered row from the others, which also helps not accidentally to lose track of the row data.


\section{Current preliminary findings}

Our preliminary experiments found patterns emerging in three different clustering cases.
In our first case, see~\autoref{fig:overview} (U1) settings in the bottom, we start to explore the test data using the attributions as clustering data and the Ward method~\cite{wardjr_hierarchical_1963} with the normalized Euclidean distance~\cite{vanwijk_cluster_1999}.
By clustering the attributions, we hypothesize that we can find patterns in the activations and raw time series.
If we inspect the visualization of~\autoref{fig:overview} (U1), we see that based on patterns we found in the attributions on the right, we also see patterns in the raw time series on the left.
Not only do we see that trend once, but on multiple occasions.
Thus, our clustering reveals subsequences in the raw data relevant to the classification.

However, now the question arises if we can also achieve such patterns using the raw time series for the clustering.
In~\autoref{fig:finding} (U2), we apply the Ward and normalized Euclidean clustering method to the raw time series.
For~\autoref{fig:finding} (U2a), our time series visualization on the left present improved patterns and better clusters, but on the attributions on the right, the patterns we found before are not or only slightly visible anymore.
We can find some previous patterns in~\autoref{fig:finding} (U2b) for the raw time series and the attributions.
However, in comparison, to~\autoref{fig:overview} (U1), the patterns of~\autoref{fig:finding} (U2b) are not as prominent.
Another clustering internally in the groups with the attributions can potentially improve the visual patterns.

After mostly neglecting them, we focus on the other visualized data, the activations.
In~\autoref{fig:finding} (U3), we can inspect the clustering results on the activations using the Ward method~\cite{wardjr_hierarchical_1963} with the normalized Euclidean distance~\cite{vanwijk_cluster_1999}.
With this data type for the clustering, we get not only some clusters in the raw time series but also noise patterns in the result for the raw time series, see~\autoref{fig:finding} (U3a).
Another clustering on top of the currently selected one can potentially support revealing other patterns in the data.
~\autoref{fig:finding} (U3b), on the other hand, presents patterns in the raw time series again, together with a few in the attributions.

\section{Conclusions and Future Work}
We presented DAVOTS, an interactive visual analytics approach to explore deep learning models trained on raw univariate time series in a dense pixel visualization.
The approach presents the input data, the activations of a specific layer, and the attributions as dense pixel displays.
We generate an ordering for the dense pixel visualization through hierarchical clustering on the different data types.
We demonstrated the approach on a popular time series benchmark dataset and a simple convolution neural network to present revealed patterns we found by an initial exploration using DAVOTS.
However, we want to extend the approach to enable further analysis.

\noindent\textbf{Future work --}
To enhance the approach, additional interactivity can help further analysis of the model and its predictions.
Some help for users can be an interactive aggregation of similar rows, e.g., by a cluster.
A drag and drop of rows to reorder the samples facilitates to find of groups and group patterns by users.
Switching the data types can also help to find cross patterns between different data attributes.
Not only can interactive aggregation support finding patterns, but an automatic transformation of the data and clustering afterward can lead to good clustering results.
For instance, we can apply a Fourier transformation to the data and then use DBSCAN to get clusters.
We further want to extend the hierarchical clustering approach with density-based and partition-based clustering and measure the results based on Guo and Gahegan~\cite{guo_spatial_2006} or a more focused measure.
Also, we look forward to applying other clustering approaches using the shape characteristics, e.g., Wang et al.~\cite{wang_characteristic_2006}, to make patterns more evident.

\section*{Acknowledgement}
This work has been partially supported by the Federal Ministry of Education and Research (BMBF) in VIKING (13N16242).



\bibliographystyle{eg-alpha-doi}

\bibliography{egbibsample}

\newcommand{\etalchar}[1]{$^{#1}$}
\begin{thebibliography}{\uppercase{VWVS99}}

\bibitem[AMST11]{aigner_visualization_2011}
\textsc{Aigner W., Miksch S., Schumann H., Tominski C.}:
\newblock \emph{Visualization of time-oriented data}, vol.~4.
\newblock Springer, 2011.

\bibitem[AS19]{assaf_explainable_2019}
\textsc{Assaf R., Schumann A.}:
\newblock Explainable deep neural networks for multivariate time series
  predictions.
\newblock In \emph{IJCAI} (2019), pp.~6488--6490.

\bibitem[ASW15]{aghabozorgi_time_2015}
\textsc{Aghabozorgi S., Shirkhorshidi A.~S., Wah T.~Y.}:
\newblock Time-series clustering--a decade review.
\newblock \emph{Information systems 53} (2015), 16--38.

\bibitem[DKK{\etalchar{*}}18]{dau_ucr_2018}
\textsc{Dau H.~A., Keogh E., Kamgar K., Yeh C.-C.~M., Zhu Y., Gharghabi S.,
  Ratanamahatana C.~A., {Yanping}, Hu B., Begum N., Bagnall A., Mueen A.,
  Batista G.}:
\newblock The {{UCR Time Series Classification Archive}}.
\newblock www.cs.ucr.edu/\textasciitilde eamonn/time\_series\_data/, Oct. 2018.

\bibitem[GG06]{guo_spatial_2006}
\textsc{Guo D., Gahegan M.}:
\newblock Spatial ordering and encoding for geographic data mining and
  visualization.
\newblock \emph{Journal of Intelligent Information Systems 27} (2006),
  243--266.

\bibitem[GMR{\etalchar{*}}18]{guidotti_survey_2018}
\textsc{Guidotti R., Monreale A., Ruggieri S., Turini F., Giannotti F.,
  Pedreschi D.}:
\newblock A survey of methods for explaining black box models.
\newblock \emph{ACM computing surveys (CSUR) 51}, 5 (2018), 1--42.

\bibitem[JNC{\etalchar{*}}20]{jeyakumar_how_2020}
\textsc{Jeyakumar J.~V., Noor J., Cheng Y.-H., Garcia L., Srivastava M.}:
\newblock How can i explain this to you? an empirical study of deep neural
  network explanation methods.
\newblock \emph{Advances in Neural Information Processing Systems 33} (2020).

\bibitem[Kei00]{keim_designing_2000}
\textsc{Keim D.~A.}:
\newblock Designing pixel-oriented visualization techniques: Theory and
  applications.
\newblock \emph{IEEE Transactions on visualization and computer graphics 6}, 1
  (2000), 59--78.

\bibitem[LL17]{lundberg_unified_2017}
\textsc{Lundberg S., Lee S.-I.}:
\newblock A {{Unified Approach}} to {{Interpreting Model Predictions}}.
\newblock In \emph{Advances in {{Neural Information Processing Systems}}}
  (2017).
\newblock \href {http://dx.doi.org/10.3321/j.issn:0529-6579.2007.z1.029}
  {\path{doi:10.3321/j.issn:0529-6579.2007.z1.029}}.

\bibitem[RSG16]{ribeiro_why_2016}
\textsc{Ribeiro M.~T., Singh S., Guestrin C.}:
\newblock "{{Why Should I Trust You}}?".
\newblock In \emph{International {{Conference}} on {{Knowledge Discovery}} and
  {{Data Mining}}} (2016).
\newblock \href {http://dx.doi.org/10.1145/2939672.2939778}
  {\path{doi:10.1145/2939672.2939778}}.

\bibitem[SAE{\etalchar{*}}19]{schlegel_towards_2019}
\textsc{Schlegel U., Arnout H., {El-Assady} M., Oelke D., Keim D.~A.}:
\newblock Towards a {{Rigorous Evaluation}} of {{XAI Methods}} on {{Time
  Series}}.
\newblock In \emph{{{ICCV Workshop}} on {{Interpreting}} and {{Explaining
  Visual Artificial Intelligence Models}}} (2019).

\bibitem[Shn96]{shneiderman_eyes_1996}
\textsc{Shneiderman B.}:
\newblock The eyes have it: a task by data type taxonomy for information
  visualizations.
\newblock In \emph{Proceedings 1996 IEEE Symposium on Visual Languages} (1996),
  pp.~336--343.
\newblock \href {http://dx.doi.org/10.1109/VL.1996.545307}
  {\path{doi:10.1109/VL.1996.545307}}.

\bibitem[SK21]{schlegel_time_2021}
\textsc{Schlegel U., Keim D.~A.}:
\newblock Time series model attribution visualizations as explanations.
\newblock In \emph{TREX: Workshop on TRust and EXpertise in Visual Analytics}
  (2021).

\bibitem[SSSE19]{spinner_explainer_2019}
\textsc{Spinner T., Schlegel U., Sch{\"a}fer H., {El-Assady} M.}:
\newblock {{explAIner}}: {{A Visual Analytics Framework}} for {{Interactive}}
  and {{Explainable Machine Learning}}.
\newblock \emph{IEEE Transactions on Visualization and Computer Graphics}
  (2019).

\bibitem[TSSG22]{theissler_explainable_2022}
\textsc{Theissler A., Spinnato F., Schlegel U., Guidotti R.}:
\newblock Explainable ai for time series classification: A review, taxonomy and
  research directions.
\newblock \emph{IEEE Access 1} (Sep 2022).
\newblock \href {http://dx.doi.org/10.1109/ACCESS.2022.3207765}
  {\path{doi:10.1109/ACCESS.2022.3207765}}.

\bibitem[VWVS99]{vanwijk_cluster_1999}
\textsc{Van~Wijk J.~J., Van~Selow E.~R.}:
\newblock Cluster and calendar based visualization of time series data.
\newblock In \emph{Proceedings 1999 IEEE Symposium on Information Visualization
  (InfoVis' 99)} (1999), IEEE, pp.~4--9.

\bibitem[WJ63]{wardjr_hierarchical_1963}
\textsc{Ward~Jr J.~H.}:
\newblock Hierarchical grouping to optimize an objective function.
\newblock \emph{Journal of the American statistical association 58}, 301
  (1963), 236--244.

\bibitem[WSH06]{wang_characteristic_2006}
\textsc{Wang X., Smith K., Hyndman R.}:
\newblock Characteristic-based clustering for time series data.
\newblock \emph{Data mining and knowledge Discovery 13} (2006), 335--364.

\end{thebibliography}


\end{document}